\documentclass[a4paper,conference]{IEEEtran}
\IEEEoverridecommandlockouts

\usepackage{cite}
\usepackage{amsmath,amssymb,amsfonts}
\usepackage{graphicx}
\usepackage{booktabs}
\usepackage{url}
\usepackage{algorithmic}
\usepackage{subcaption}
\usepackage[normalem]{ulem}

\newcommand{\authorfont}{\fontsize{11pt}{10pt}\selectfont}
\renewcommand\IEEEkeywordsname{Keywords}

\usepackage[T1]{fontenc}
\usepackage[latin9]{inputenc}
\usepackage{babel}
\usepackage[table]{xcolor}
\usepackage{collcell}
\usepackage{hhline}
\usepackage{pgf}
\usepackage{multirow}

\usepackage[table]{xcolor}
\usepackage{hhline}
\usepackage{pgf}
\usepackage{tikz}
\usepackage{collcell}
\usepackage{bbding} 
\definecolor{Gray}{gray}{0.9}

\usepackage{adjustbox}
\usepackage{array}

\usepackage{ifthen}
\usetikzlibrary{matrix,calc}

\newcommand{\bftab}{\fontseries{b}\selectfont}
\usepackage{booktabs}
\usetikzlibrary{positioning}
\usepackage{multicol}
\usepackage[hidelinks]{hyperref}
\begin{document}
\bstctlcite{IEEEexample:BSTcontrol}

\title{Grammar Based Speaker Role Identification for Air Traffic Control Speech Recognition}

\author{\IEEEauthorblockN{\authorfont Amrutha Prasad$^{\star,1,2}$, Juan Zuluaga-Gomez$^{1,3}$, Petr Motlicek$^{1,2}$, Saeed Sarfjoo$^{1}$, \\ Iuliia Nigmatulina$^{1,4}$, Oliver Ohneiser$^{5}$, Hartmut Helmke$^{5}$
}
\IEEEauthorblockA{
\\ \normalsize $^{1}$Idiap Research Institute, Martigny, Switzerland\\
\normalsize $^{2}$Brno University of Technology, Speech@FIT, IT4I CoE, Brno, Czech Republic \\
\normalsize $^{3}$Ecole Polytechnique Federale de Lausanne (EPFL), Lausanne, Switzerland \\
\normalsize $^{4}$Institute of Computational Linguistics, University of Zurich, Switzerland \\
\normalsize $^{5}$German Aerospace Center (DLR), Institute of Flight Guidance, Braunschweig, Germany \\
\normalsize $^\star$corresponding author: amrutha.prasad@idiap.ch
}
}

\maketitle
\newcommand{\xvec}{\mathbf{x}}
\newcommand{\observation}{\mathbf{o}}
\newcommand{\wvec}{\mathbf{w}}
\newcommand{\model}{\mathcal{M}}
\newcommand{\mden}{\model_{\mathrm{den}}}
\newcommand{\mmi}{\mathcal{F}}
\newcommand{\enn}{\mathcal{t}}
\newcommand{\costfunc}{\mmi_{\text{MMI}}}
\newcommand{\costfuncone}{\mmi_{\text{MMI}}^{(1)}}
\newcommand{\costfuncL}{\mmi_{\text{MMI}}^{(T)}}
\newcommand{\costfuncl}{\mmi_{\text{MMI}}^{(t)}}
\newcommand{\parameters}{\boldsymbol{\theta}}
\newcommand{\sharedparams}{\boldsymbol{\theta}_{\text{shared}}}
\newcommand{\langparams}{\boldsymbol{\theta}_{n}}
\newcommand{\langweight}{\alpha_{t}}
\newcommand{\postnum}{^{\mathrm{NUM}}\gamma}
\newcommand{\postden}{^{\mathrm{DEN}}\gamma}
\newcommand{\postnumind}{\gamma^{\mathrm{num-ind}}}
\newcommand{\postdenind}{\gamma^{\mathrm{den-ind}}}
\newcommand{\postnumdep}{^{\mathrm{NUM}_{t}}\gamma}
\newcommand{\postdendep}{^{\mathrm{DEN}_{t}}\gamma}
\newcommand{\subl}{t}

\tikzset{%
  every neuron/.style={
    circle,
    draw,
    minimum size=0.5cm
  },
  neuron missing/.style={
    draw=none, 
    scale=3,
    execute at begin node=\color{black}$\hdots$
  },
}

\tikzstyle{block} = [draw, rectangle, 
    minimum height=3em, minimum width=7em]

\newcommand\blfootnote[1]{%
  \begingroup
  \renewcommand\thefootnote{}\footnote{#1}%
  \addtocounter{footnote}{-1}%
  \endgroup
}
\begin{abstract}
Automatic Speech Recognition (ASR) for air traffic control is generally trained by pooling Air Traffic Controller (ATCO) and pilot data into one set. 
This is motivated by the fact that pilot's voice communications are more scarce than ATCOs.
Due to this data imbalance and other reasons (e.g., varying acoustic conditions), the speech from ATCOs is usually recognized more accurately than from pilots. 
Automatically identifying the speaker roles is a challenging task, especially in the case of the noisy voice recordings collected using Very High Frequency (VHF) receivers or due to the unavailability of the push-to-talk (PTT) signal, i.e., both audio channels are mixed. In this work, we propose to (1) automatically segment the ATCO and pilot data based on an intuitive approach exploiting ASR transcripts and (2) subsequently consider an automatic recognition of ATCOs' and pilots' voice as two separate tasks. Our work is performed on VHF audio data with high noise levels, i.e., signal-to-noise (SNR) ratios below 15 dB,
as this data is recognized to be helpful for various speech-based machine-learning tasks. Specifically, for the speaker role identification task, the module is represented by a simple yet efficient knowledge-based system exploiting a grammar defined by the International Civil Aviation Organization (ICAO). The system accepts text as the input, either manually verified annotations or automatically generated transcripts. The developed approach provides an average accuracy in speaker role identification of about 83\%. Finally, we show that training an acoustic model for ASR tasks separately (i.e., separate models for ATCOs and pilots) or using a multitask approach is well suited for the noisy data and outperforms the traditional ASR system where all data is pooled together. 
\end{abstract}



\section{Introduction}
\label{sec:introduction}

Previous work~\cite{srinivasamurthy2017semi,kleinert2018semi} as part of the MALORCA\footnote{MAchine Learning Of speech Recognition models for Controller Assistance: \url{http://www.malorca-project.de/wp/} } and AcListant-Strips\footnote{Active Listening Assistant Strips: \url{https://www.malorca-project.de/wp/?page_id=350} } projects, focused on i) improving Assitant-Based Speech Recognition (ABSR) accuracy only for ATCOs, ii) reducing workload for ATCOs~\cite{helmke2016reducing}, and iii) increasing  efficiency~\cite{helmke2017increasing} of ATCOs.
In the ongoing HAAWAII\footnote{Highly Advanced Air Traffic Controller Workstation with Artificial Intelligence Integration: \url{https://www.haawaii.de}} project, research focuses on developing a reliable and adaptable solution to transcribe voice commands issued by both ATCOs and pilots automatically.


An error-resilient and accurate ASR system is critical in the ATC domain.
Current state-of-the-art technologies require large amounts of data to train ASR systems. The goal of a recently finished project called ATCO2\footnote{Automatic collection and processing of voice data from air-traffic communications \url{https://www.atco2.org/}.} was to collect and automatically transcribe a large database of voice recordings of ATCOs and pilots (with a minimum effort) for the purpose mentioned above~\cite{kocour21_interspeech,zuluaga2020automatic,rigault2022legal}. 
ATCO and pilot speech recordings are usually pooled together~\cite{zuluagagomez20_interspeech,zuluaga2020automatic,srinivasamurthy2017semi} to train the ASR despite having a significant variability in the data distribution (acoustic and grammatical conditions), and the number of speakers in the data.
As a result of the variability in the data distribution, ASR performances are significantly different for ATCO and pilot speech.\footnote{The air traffic controllers' speech is more straightforward to recognize than pilots'.} Our baseline system trained by pooling all data and evaluated on noisy data (signal-to-noise ratio (SNR) below 15 dB) shows an absolute difference in word error rate (WER) of 9.7\% on ATCO and pilot speech (ATCO WER: 36.1\%, Pilot WER: 45.8\%). ASR on another dataset also revealed that it is 'twice as hard ' to correctly recognize pilot utterances compared to ATCO utterances due to shortened speech~\cite{pellegrini2020atcopilotasr}.

In this paper, we hypothesize that introducing information about the speaker role during ASR training can help mitigate this variability. We consider the approach of training the acoustic model in the ASR to produce outputs for each speaker role--ATCO and pilot--separately. This is often called multitask learning in Deep Neural Networks parlance~\cite{ruder2017overview}. Specifically, this paper investigates a multitask approach to training AMs to be integrated into ASR for ATCO and pilot. The obvious first step is automatically splitting the ATC speech communication data into two speaker roles. However, obtaining speaker labels manually on a large dataset is expensive and time-consuming. 
A common approach is to diarize the audio~\cite{anguera2012speaker,park2021review} (see Section~\ref{sec:related-work-spk}). Although the ATCO speech is often cleaner (higher SNR value) than the pilot (as the former communicates from a controlled acoustic environment), the speech recordings collected in ATCO2 project using Very High Frequency (VHF) receivers\footnote{Blog related to setting up receivers \url{https://www.atco2.org/news/setting-up-vhf-receiver-for-air-traffic-communication}} are noisy for both ATCO and pilot channels~\cite{zuluaga2022atco2}. In such a case, a speaker diarization system may fail to assign speaker labels accurately. Thus, it is not advisable to rely fully on a pure acoustic-based system to obtain accurate speaker labels.



Another approach to obtain the speaker class is through leveraging the `ICAO' grammar to classify an utterance as one of the classes based on text.
The ICAO grammar~\cite{allclear} is a well-defined, standard phraseology to ensure safe communication between a controller and pilot, which in turns assure smooth travel of the aircraft. 
Once the speaker role labels are available over a large database, AMs can be trained for both ATCOs and pilots with different approaches. In this study, we show that due to the poor acoustic conditions, training a single acoustic model (AM) by pooling all data can result in degradation of the ASR performance on pilot speech.
To obtain better accuracy, AM should be trained separately for ATCO and pilot data, or considered as different tasks by using a multitask approach.

Our paper is structured as follows. Section 2 provides a brief overview of the work related to multitask automatic speech recognition. The datasets used are described in Section 3 followed by Section 4 that describes speaker role classification with text.
Section 5 explains the experimental setup and the results obtained, which are followed by the conclusion in Section 6. 

\begin{figure}[t]
  \centering
  \includegraphics[width=0.35\textwidth]{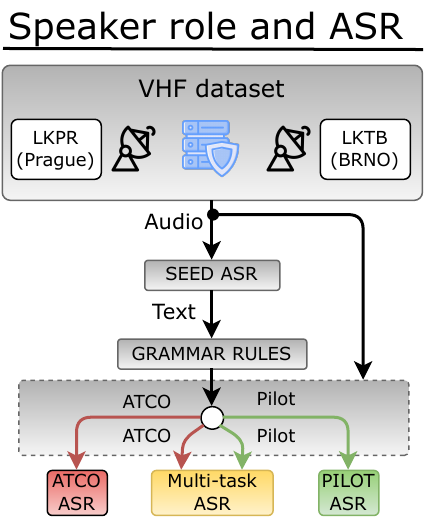}
  \caption{\textbf{Overall pipeline to train a multitask ABSR system}. A seed ABSR is used to transcribe all the available VHF data. Then we use the grammar-based module to split the databases by speaker roles, i.e., ATCO or pilot. Then, the split databases are used to train speaker-dependent acoustic models in the case of separate ATCO and pilot models. The same information is also used by the multitask system to select the task to be trained for each utterance. The same procedure is applied to other datasets used in this paper.} 
  \label{fig:grammar-based-sid}
\end{figure}

\section{Related Work}

\subsection{Speaker Role Classification}
\label{sec:related-work-spk}

The current approach to identify a speaker role for a given utterance is extracted from an acoustic-based diarization system.
A speaker diarization system can be defined as a task of defining speaker roles to segments of an utterance. This approach is developed over the years for performing segmentation of conversations or for finding a speaker of interest in a given set of speakers. Several approaches based on Bayesian hidden Markov model and neural networks have evolved over the past two decades, which are current state-of-the-art.

Communication between ATCOs and pilots is collected as a single channel, typically without VAD; thus, the obtained audio files end spamming several minutes. Therefore, the first step conveys to apply segmentation, i.e., typically a Voice Activity Detection (VAD) system. This system splits the audio based on long silence regions to get a set of audio files to which diarization can then be applied. 
The current acoustic-based speaker role classification system used in HAAWAII employs the clustering of speaker embeddings (x-vectors) approach described by Landini et al.~\cite{landini2022bayesian}. Firstly, a neural network is trained to discriminate between various speakers, so embeddings can capture the relevant information, which allows comparing speech fragments and deciding if they belong to the same speaker. The output of this neural network is the x-vectors. The second step is clustering these x-vectors based on the Bayesian hidden Markov model, where each state in the model is represented as one speaker. When finding the state that most likely is produced by a given x-vector, the x-vector is assigned one speaker. The final diarization output is the assignment of an x-vector to a speaker role. 

\begin{figure*}[t]
  \centering
  \includegraphics[width=0.95\textwidth]{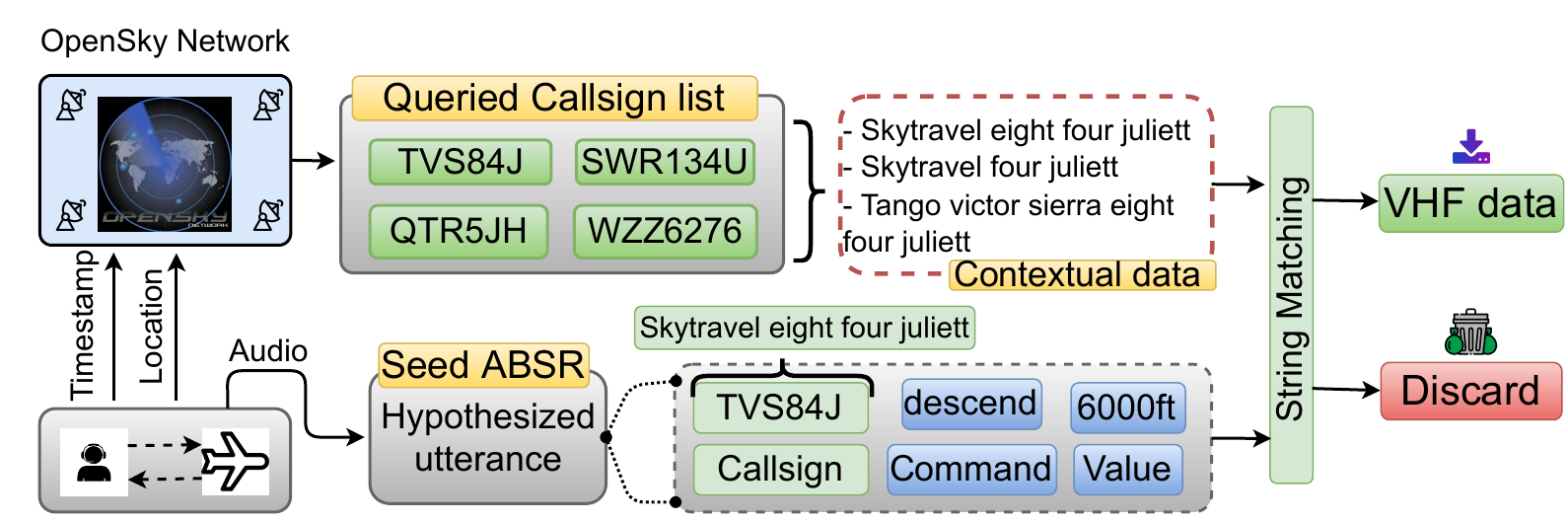}
  \caption{\textbf{Pipeline for gathering ATCO-pilot voice communications with VHF receivers}. Timestamp and location of the communication is used to query a callsign list (surveillance data) from OpenSky Network servers. In parallel, the communication is transcribed with an in-domain ABSR system. Later, the contextual data is verbalized, i.e., ICAO callsign $\rightarrow$ spoken words (red doted box). Finally, a string matching algorithm iteratively tries to match the callsigns in the surveillance data with the words hypothesized by the ABSR, i.e., the transcripts. If a match is found, the segment is stored, otherwise, it is discarded.} 
  \label{fig:osn_vhf_data}
\end{figure*}

In a typical recording of ATC communications, one ATCO would communicate with
several pilots. Hence, most parts of these speeches are marked as ATCOs and the rest as pilots based on the former prior assumption. Since our primary goal is to identify ATCO/pilot, many speakers share the same speaker role label, and thus a speaker diarization system will not be easy to train. This makes the clustering and the final stage of deciding the speaker role label complicated, and the diarization system might fail. That is one of the main reasons behind our idea to develop a grammar-based speaker role detection based on text (e.g., ASR transcripts).

\subsection{Multi-task ASR}
\label{sec:related-work-asr}

Previous research has shown that to compensate for limited data available in low-resourced languages, multilingual systems are an effective way to train ASR systems~\cite{madikeri2020lattice,burget2010multilingual,imseng2014using,vu2014multilingual,karafiat2016multilingual}. 
In such a system, the output layer could be a separate layer for each language, or a single layer shared between all languages~\cite{karafiat2016multilingual}. The Kaldi~\cite{povey2011kaldi} toolkit provides state-of-the-art techniques to train ABSR, specifically, Lattice-Free Maximum Mutual Information (LF-MMI) based models~\cite{povey2016purely}. 
Recently,~\cite{madikeri2020lattice} showed that multilingual AM can be trained with LF-MMI~\cite{povey2016purely}.
In MMI training, the cost function is given as:
\begin{equation}
\mmi _{\text{MMI}} = \sum_{u=1}^{U} \log 
\frac{p\left(\xvec^{(u)} \vert \model_{\wvec(u)}, \parameters \right) p(\wvec(u))}
{p\left(\xvec^{(u)} \vert \mden, \parameters \right)} ,
\label{eq:mmi}
\end{equation}
\noindent where 
$\xvec^{(u)}$ is an input sequence for an utterance $u$,
$U$ is a set of all utterances in the training data, 
$\model_{\wvec(u)}$ corresponds to a numerator graph specific to a word sequence in transcription, 
$\mden$ is a denominator graph modelling all possible sequences which is usually a phone LM,
$\parameters$ is a model parameter and
$p(\wvec(u))$ is a language model probability for an utterance.

However, in multitask training with separate output layers, the cost function from Equation~\ref{eq:mmi} is computed for each task depending on the number of tasks. For $T$ tasks, the output cost function for each task $t$ depends only on the utterances of that task:
\begin{equation}
\costfuncl = \sum_{u=1}^{U_{\subl}} \log \frac{p\left(\xvec^{(u)} \vert \model_{\wvec(u)}^{t}, \parameters \right) p(\wvec(u))}
{p\left(\xvec^{(u)} \vert \mden^{t}, \parameters \right)} ,
\end{equation}
where $U_{t}$ is the number of utterances in a minibatch for a task $t$, $\parameters$ contains the shared and task-dependent parameters, $\model_{\wvec(u)}^{t}$ and $\mden^{t}$ are task-specific numerator and denominator graphs, respectively.
For a task $t$, a denominator graph is built using the task-specific phone.
For each minibatch, the gradient of each task output layer is computed and updated.

The overall cost-function is then given as a weighted sum of all task-dependent cost-functions defined in Equation~\ref{eq:finalml}.

\begin{equation}
\costfunc = \sum_{t=1}^{T} \langweight  {F}^{t}_{MMI},
\label{eq:finalml}
\end{equation}
\noindent where $\langweight$ is a task-dependent weight.

Although language and phone sets are the same for ATCO and pilots, due to the variation in the acoustic conditions, we consider them as different tasks and propose to use a multitask approach to train AMs. We hypothesize that using a multitask approach can lead to  better ASR performance for both ATCOs and pilots compared to a single AM trained by combining all data.

Different approaches to improve the ASR performance have explored semi-supervised learning~\cite{srinivasamurthy2017semi,kleinert2018semi,zuluagagomez21_interspeech}, integration of surveillance data as prior knowledge into the ASR pipeline~\cite{kocour2021automatic,nigmatulina2021improving,nigmatulina2022two,kocour21_interspeech} and end-to-end training~\cite{zuluaga2022does} for ATC domain. Additionally, related work on text-based diarization for ATC communications is explored in~\cite{zuluaga2021bertraffic}. 


\section{Datasets}
\label{sec:datasets}

\begin{table}[t]
    \caption{\textbf{Air traffic control communications-related databases used for training. The whole database of HAAWAII and ATCO2 were not available during this work. } $^\dagger$total number of audio in the database after silence removal.}
    \label{tab:databases}
    \centering
    \begin{tabular}{cccccc}
        \toprule
        \rowcolor{Gray} \textbf{Database} &  & \textbf{Duration}$^\dagger$ &  & \textbf{Open} & \textbf{Ref} \\
        \rule{0pt}{2ex}  &  & Training &  & \textbf{source}  &  \\
        \midrule
        \rowcolor{Gray} \multicolumn{6}{c}{\textbf{Private databases}} \\
        \midrule
        HAAWAII &  & 43 &  & \XSolidBrush &  \cite{zuluaga2022does} \\
        MALORCA & & 13 & & \XSolidBrush & \cite{kleinert2018semi,srinivasamurthy2017semi}  \\
        AIRBUS & & 100 & & \XSolidBrush &  \cite{AIRBUS} \\
        \midrule
        \rowcolor{Gray} \multicolumn{6}{c}{\textbf{Public databases}} \\
        \midrule
        ATCOSIM &  & 8  &  & \checkmark & \cite{ATCOSIM} \\
        UWB-ATCC &  & 10.4  &  & \checkmark & \cite{UWB_ATCC} \\
        LDC-ATCC &  & 23  &  & \checkmark & \cite{LDC_ATCC} \\
        HIWIRE & & 28.7  & & \checkmark &  \cite{HIWIRE} \\
        ATCO2 &  & 5000 &  & \checkmark & \cite{kocour2021automatic} \\
        \bottomrule
    \end{tabular}
\end{table}

The following subsections provide an overview of the public and private databases used in this paper. A brief overview is also provided in Table~\ref{tab:databases}. 
\subsection{Collection and Pre-processing of VHF Data}
\label{sec:vhfdata}

\subsubsection{Data collection}

To obtain ATC voice communications, the following two sources are considered: (i) open-source speech like LiveATC,\footnote{LiveATC.net is a streaming audio network consisting of local receivers tuned to aircraft communications: \url{https://www.liveatc.net/}.} and ii) speech collected  with our own setup of VHF receivers. 
In addition to speech data, the time-aligned metadata available is used to obtain surveillance data (e.g., callsign list for each communication) from OpenSky Network\footnote{
OpenSky Network is a non-profit association based in Switzerland. It aims at improving the security, reliability, and efficiency of the airspace usage by providing open access of real-world air traffic control data to the public. The OpenSky Network consists of a multitude of sensors connected to the Internet by volunteers, industrial supporters, and academic/governmental organizations. URL: \url{https://opensky-network.org}.} (OSN). 
This iterative process yielded $\sim$377 hours of speech data from Prague (LKPR) and Brno (LKTB) airports from August 2020 until January 2021. 
This subset is part of a full corpus of around five thousand hours of ATC audio and metadata collected during the ATCO2 project. The full corpus is available for purchase through ELDA in \url{http://catalog.elra.info/en-us/repository/browse/ELRA-S0484}. The recordings of both corpus are mono-channel sampled at 16kHz and 16-bit PCM.
In this paper, the whole corpus of five thousand hours is not used, as it wasn't available at the time of experimentation.





\subsubsection{Data pre-processing}
\label{sec:data-pre-processing}

Figure \ref{fig:osn_vhf_data} shows the pipeline used for preparing the VHF database. 
First, a seed ASR system is used to produce the transcripts for the 377 hours of collected data. The seed model is a `hybrid' speech-to-text recognizer based on Kaldi~\cite{povey2011kaldi} trained with the LF-MMI cost function~\cite{povey2016purely} (see Section~\ref{sec:related-work-asr}).
The neural network follows a {\tt cnn-tdnn} topology. It has six \texttt{convolutional-layers} followed by nine layers of Factorized Time-Delay Neural Network (\texttt{TDNN-F})~\cite{povey2018semi}.


A list of callsigns is retrieved from OSN in ICAO format. Callsigns in ICAO format are composed of three characters airline code, e.g., \textit{TVS}, followed by a flight number which can consist of digits or letters, e.g., \textit{TVS84J}. In order to use this prior knowledge in the ASR, we first verbalize the ICAO callsigns, i.e., transform to the ``expanded representation''. Several variants exist for a given callsign. 
As illustrated in Figure~\ref{fig:osn_vhf_data}, the callsign TVS84J can be pronounced as \textcolor{blue}{``\dashuline{skytravel eight four juliett}"} or instead each letter can be spelled out as \textcolor{blue}{``\dashuline{tango victor sierra eight four juliett}"}. For more information about these rules, see~\cite{kocour2021automatic}. Then, an expanded list of callsigns with its variants is created. Finally, string matching of this expanded callsign list is applied to the transcripts generated by the ABSR system. 
The utterances in which one of the callsigns is found are stored, while the rest are discarded. This pre-processing reduced the data from 377 hours to 66 hours.

\begin{figure}[t]
  \centering
  \includegraphics[width=0.32\textwidth]{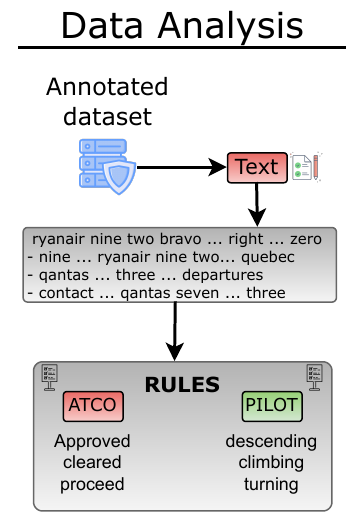}
  \caption{\textbf{Overall process to develop a rule-based grammar system to identify ATCOs and pilots based on text}. We perform a data analysis on the already available public and private databases. We then gather a set of words that are more probable to appear in either, ATCO or pilot utterances.} 
  \label{fig:data_analysis}
\end{figure}

\subsection{Related ATC Datasets Available for Training}
\label{OthData}

In addition to the above data collection, ATCO2 has brought together several air traffic command-related databases 
~\cite{srinivasamurthy2017semi,PILSEN_ATC,N4NATO,HIWIRE,ATCOSIM,AIRBUS,LDC_ATCC} from different publicly available open data sources. The full set of databases span approximately 100 hours of speech data that are strongly related in both, phraseology and structure seen in ATCO-pilot communications~\cite{zuluagagomez20_interspeech,zuluaga2020automatic,zuluagagomez21_interspeech}. The collection of databases is augmented by adding noise that matched LiveATC audio channels, doubling the size of training data. Additionally, since each of the seven databases had different annotation ontologies (annotation procedure, rules, and symbols), the transcripts had to be standardized and normalized~\cite{ATCOSIM,helmke2018ontology}.

\section{Speaker Role Classification with Text}
\label{sec:spkclass}

As described in Section~\ref{sec:introduction}, to develop a reliable and improved ASR for both, ATCOs and pilots, respective labelled speech data are required. However, in most cases (e.g., ATCO2 project) although large amounts of data are collected, they do not contain speaker labels. The first task is therefore to split the speech recordings into two classes: ATCO and pilot. To accomplish this, we extract the information based on the ICAO grammar to identify the speaker's role. We follow the pipeline described in Figure~\ref{fig:data_analysis}.

\begin{table}[t]
    \centering
    \caption{\textbf{List of ATCO and Pilot words used in the proposed grammar-based classification system}. We collected 31 words for ATCO and 20 words for pilot.}
    \label{tab:atcopilotwords}
    \begin{tabular}{@{}cccc@{}}
        \toprule
        \rowcolor{Gray} \multicolumn{4}{c}{\textbf{ATCO words}} \\
        \midrule
        approved & back & break & call \\
        cleared & contact & correct & direct \\
        disregard & established & expect & handover \\
        identified & increase & maintain & no \\
        proceed & radar & reduce & report \\
        roger & soon & standby & transition \\
        turn & vortex & wake & wind \\
        you're & you've & yours & \\
        \midrule
        \rowcolor{Gray} \multicolumn{4}{c}{\textbf{Pilot words}}\\
        \midrule
        CPDLC & approaching & climbing & comply \\
        descending & heavy & inbound & maintaining \\
        our & reducing & request & requesting \\
        standing & stopping & taking & turning \\
        us & we & will & wilco \\
        \bottomrule
    \end{tabular}
\end{table}

ICAO defines a separate grammar for ATCOs and pilots to enable clear communication. For instance, there are certain phrases/commands that an ATCO should use in specific order. This knowledge is used to extract/identify potential words/commands that indicate a specific role of speaker.
For example, the words such as ``\texttt{identified}", ``\texttt{approved}", ``\texttt{wind}" would most probably only be spoken by an ATCO and the words ``\texttt{wilco}", ``\texttt{maintaining}", ``\texttt{we}", ``\texttt{our}" would probably be spoken only by a pilot.
Currently, we have made a list of 31 words for ATCO and 20 words for pilot that indicate each role. The list of words is presented in Table~\ref{tab:atcopilotwords}, while the overall pipeline to gather these words is depicted in Figure~\ref{fig:data_analysis}.
This list was generated by manual curation and expert feedback. A list of callsigns\footnote{The table lists the IATA airline designators, the ICAO airline designators and the airline callsigns (telephony designator). URL: \url{https://en.wikipedia.org/wiki/List_of_airline_codes}.} is also prepared from available airline codes.

Since this method operates at word level, manual (if available) or automatically generated transcripts are required for the corresponding speech recordings. In order to identify if an utterance is spoken by an ATCO or a pilot, we check the corresponding transcript for the conditions below: if the callsign appears at the beginning of an utterance, this utterance is classified as ATCO, else it is classified as a pilot. As there is a greeting at the beginning quite often, we check if the callsign appears within the first four words. If one of the words in the utterance is in the list of ATCO words or in the list of pilot words, then the respective role is assigned.

Once each utterance in the training data is tagged with ATCO or pilot labels, we propose to train two versions of ASR. In the first system, there are two acoustic
models: one for ATCO and one for pilot. In the second system, we train a multitask network with one task as ATCO ASR and the other as pilot ASR (see Section~\ref{sec:related-work-asr}). The procedure is illustrated in Figure~\ref{fig:grammar-based-sid}.




\def\myConfMat{{
{338,   78 },  
{53,  397 },  
}}
\def\myConfMatTwo{{
{435,   133 },  
{65,  470 },  
}}

\def\myConfMatThree{{
{588,   288 },  
{193,  699 },  
}}

\def\classNames{{"ATCO","Pilot"}} 

\def\numClasses{2} 

\def\myScale{1.5} 

\begin{figure}[t]
    \centering
    \begin{tikzpicture}[scale = \myScale,]
        \tikzset{vertical label/.style={rotate=90,anchor=east}}   
        \tikzset{diagonal label/.style={rotate=45,anchor=north east}}
        
        \foreach \y in {1,...,\numClasses} 
        {
        \node [anchor=east] at (0.4,-\y) {\pgfmathparse{\classNames[\y-1]}\pgfmathresult}; 
        
        \foreach \x in {1,...,\numClasses}  
        {
        \def\totSamples{0}
        \foreach \ll in {1,...,\numClasses}
        {
            \pgfmathparse{\myConfMat[\ll-1][\x-1]}   
            \xdef\totSamples{\totSamples+\pgfmathresult} 
        }
        \pgfmathparse{\totSamples} \xdef\totSamples{\pgfmathresult}  
        
        \begin{scope}[shift={(\x,-\y)}]
            \def\mVal{\myConfMat[\y-1][\x-1]} 
            \pgfmathtruncatemacro{\r}{\mVal}   %
            \pgfmathtruncatemacro{\p}{round(\r/\totSamples*100)}
            \coordinate (C) at (0,0);
            \ifthenelse{\p<50}{\def\txtcol{black}}{\def\txtcol{white}} 
            \node[
                draw,                 
                text=\txtcol,         
                align=center,         
                fill=black!\p,        
                minimum size=\myScale*10mm,    
                inner sep=0,          
                ] (C) {\r\\\p\%};     
            \ifthenelse{\y=\numClasses}{
            \node [] at ($(C)-(0,0.75)$) 
            {\pgfmathparse{\classNames[\x-1]}\pgfmathresult};}{}
        \end{scope}
        }
        }
        \coordinate (yaxis) at (-0.5,0.3-\numClasses/2);  
        \coordinate (xaxis) at (0.5+\numClasses/2, -\numClasses-1.15); 
        \node [vertical label] at (yaxis) {Predicted Class };
        \node []               at (xaxis) {Actual };
    \end{tikzpicture}
    \caption{\textbf{Confusion matrix for speaker role identification based on text for manually speaker segmented data for London Approach test set}. The total number of ATCO utterances are 391 and the total number of pilot utterances are 475. } 
    \label{fig:CM-NATS}
\end{figure}

\begin{figure}[t]
    \centering
    \begin{tikzpicture}[scale = \myScale]
        \tikzset{vertical label/.style={rotate=90,anchor=east}}   
        \tikzset{diagonal label/.style={rotate=45,anchor=north east}}
        
        \foreach \y in {1,...,\numClasses} 
        {
        \node [anchor=east] at (0.4,-\y) {\pgfmathparse{\classNames[\y-1]}\pgfmathresult}; 
        
        \foreach \x in {1,...,\numClasses}  
        {
        \def\totSamples{0}
        \foreach \ll in {1,...,\numClasses}
        {
            \pgfmathparse{\myConfMatTwo[\ll-1][\x-1]}   
            \xdef\totSamples{\totSamples+\pgfmathresult} 
        }
        \pgfmathparse{\totSamples} \xdef\totSamples{\pgfmathresult}  
        
        \begin{scope}[shift={(\x,-\y)}]
            \def\mVal{\myConfMatTwo[\y-1][\x-1]} 
            \pgfmathtruncatemacro{\r}{\mVal}   %
            \pgfmathtruncatemacro{\p}{round(\r/\totSamples*100)}
            \coordinate (C) at (0,0);
            \ifthenelse{\p<50}{\def\txtcol{black}}{\def\txtcol{white}} 
            \node[
                draw,                 
                text=\txtcol,         
                align=center,         
                fill=black!\p,        
                minimum size=\myScale*10mm,    
                inner sep=0,          
                ] (C) {\r\\\p\%};     
            \ifthenelse{\y=\numClasses}{
            \node [] at ($(C)-(0,0.75)$) 
            {\pgfmathparse{\classNames[\x-1]}\pgfmathresult};}{}
        \end{scope}
        }
        }
        \coordinate (yaxis) at (-0.5,0.3-\numClasses/2);  
        \coordinate (xaxis) at (0.5+\numClasses/2, -\numClasses-1.15); 
        \node [vertical label] at (yaxis) {Predicted Class };
        \node []               at (xaxis) {Actual };
    \end{tikzpicture}
    \caption{\textbf{Confusion matrix for speaker role identification based on text for manually speaker segmented data for Icelandic en-route test set}. The total number of ATCO utterances are 500 and the total number of pilot utterances are 604. } 
    \label{fig:CM-Isavia}
\end{figure}

\begin{figure}[t]
    \centering
    \begin{tikzpicture}[scale = \myScale]
        \tikzset{vertical label/.style={rotate=90,anchor=east}}   
        \tikzset{diagonal label/.style={rotate=45,anchor=north east}}
        
        \foreach \y in {1,...,\numClasses} 
        {
            \node [anchor=east] at (0.4,-\y) {\pgfmathparse{\classNames[\y-1]}\pgfmathresult}; 
            
            \foreach \x in {1,...,\numClasses}  
            {
            \def\totSamples{0}
            \foreach \ll in {1,...,\numClasses}
            {
                \pgfmathparse{\myConfMatThree[\ll-1][\x-1]}   
                \xdef\totSamples{\totSamples+\pgfmathresult} 
            }
            \pgfmathparse{\totSamples} \xdef\totSamples{\pgfmathresult}  
            
            \begin{scope}[shift={(\x,-\y)}]
                \def\mVal{\myConfMatThree[\y-1][\x-1]} 
                \pgfmathtruncatemacro{\r}{\mVal}   %
                \pgfmathtruncatemacro{\p}{round(\r/\totSamples*100)}
                \coordinate (C) at (0,0);
                \ifthenelse{\p<50}{\def\txtcol{black}}{\def\txtcol{white}} 
                \node[
                    draw,                 
                    text=\txtcol,         
                    align=center,         
                    fill=black!\p,        
                    minimum size=\myScale*10mm,    
                    inner sep=0,          
                    ] (C) {\r\\\p\%};     
                \ifthenelse{\y=\numClasses}{
                \node [] at ($(C)-(0,0.75)$) 
                {\pgfmathparse{\classNames[\x-1]}\pgfmathresult};}{}
            \end{scope}
            }
        }
        \coordinate (yaxis) at (-0.5,0.3-\numClasses/2);  
        \coordinate (xaxis) at (0.5+\numClasses/2, -\numClasses-1.15); 
        \node [vertical label] at (yaxis) {Predicted Class };
        \node []               at (xaxis) {Actual };
    \end{tikzpicture}
    \caption{\textbf{Confusion matrix for speaker role identification based on text for manually speaker segmented data for LiveATC data}. The total number of ATCO utterances are 781 and the total number of pilot utterances are 987. } 
    \label{fig:CM-LiveATC}
\end{figure}

\subsection{Assigning Scores to Decisions}

The grammar role also provides the probability of assigning a speaker role to a given utterance using the bag-of-words that are manually created. In order to obtain such probability, Bayes' rule is adopted. For instance, the probability of an utterance being ATCO is computed as: 

\begin{equation}
p(\text{atco} | \text{utt}) = \frac{p(\text{utt} | \text{atco}) p(\text{atco})} {p(\text{utt} | \text{atco}) p(\text{atco}) + p(\text{utt} | \text{pilot}) p(\text{pilot})},
\end{equation}

Here $p(\text{atco})$ and $p(\text{pilot})$ are the priors, and we assume both classes have equal probability and hence their value is 0.5.
The $p(\text{utt}| \text{atco})$ is computed as:

\begin{equation}
p(\text{utt} | \text{atco}) = \prod_{w_{i} \in \text{utt}} p(w_{i} | \text{atco}).
\end{equation}

Similarly, the $p(\text{utt} | \text{pilot})$ is computed as:

\begin{equation}
p(\text{utt} | \text{pilot}) = \prod_{w_{i} \in \text{utt}}  p(w_{i} | \text{pilot}).
\end{equation}

The $p(w_{i} | \text{atco})$ and $p(w_{i} | \text{pilot})$ are computed from using the 15k speaker role annotated utterances available as part of HAAWAII project from the Air Navigation Service Providers (ANSPs) for training: i) NATS for London Approach and ii) ISAVIA for Icelandic en-route where the total number of utterances for ATCO and pilot are 7k and 8k respectively. The below equation is used to compute this:

\begin{equation}
    p(w_{i} | \text{class}) = \frac{\text{class\ count}}{\text{total\ count}}, 
\end{equation}

where $\text{class count}$ is the number of times the word $w_i$ appears in that particular class, and $\text{total count}$ is the sum of the number of times the words in both the classes.

\subsection{Speaker Role Classification Performance}

This method has been tested on manually speaker segmented and transcribed data for three different test sets: i) NATS for London Approach, ii) ISAVIA for Icelandic en-route and iii) LiveATC test set.
In the first set, there are 391 and 475 ATCO and pilot utterances, respectively. From the confusion matrix shown in Figure~\ref{fig:CM-NATS}, we can observe that this method provides a true positive rate (TPR) of $86\%$ (correctly classified ATCO) and true negative rate (TNR) of $84\%$ (correctly classified pilot). The second set used consists of 500 ATCO utterances and 604 pilot utterances. From the confusion matrix shown in Figure~\ref{fig:CM-Isavia}, we see that this method provides a TPR of $87\%$ and TNR of $78\%$. For the third set, we see a TPR of $75\%$ and a TNR of $71\%$. This shows that the bag-of-words generated match the first two sets and the communication is slightly different since there is a domain mismatch caused by data from different airports.

\subsection{Error Analysis}

As there exists many variants for any given callsign, checking only for the airline code (e.g., lufthansa) is a major factor contributing to the misclassification of ATCO as pilot. A reason for the misclassification of pilot as ATCO is the occurrence of callsigns at the beginning of the utterance. 
Analysis of misclassification errors show that the accuracy can be improved by i) matching the callsign spoken with its allowed variants\footnote{For instance, LUF189AF $\rightarrow$ lufthansa one eight nine alfa foxtrot, one eight nine alfa foxtrot, etc.} and ii) using the context prior to the callsigns.\footnote{An example is that the pilot may mention the place of the control they want to communicate followed by the callsign} We will consider applying the aforementioned improvements as a part of our future work.

\begin{table}[t]
    \caption{\textbf{Comparison of word error rates (WER) in percentages for acoustic models trained with data from other ATC datasets}. The models are tested on LiveATC ATCO and pilot test sets. The results show that training speaker-dependent acoustic models or a multitask system provide better ASR performance than the combined system.}
    \label{tab:mt1}
    \centering
    \scalebox{1.2}{
    \begin{tabular}{ ccc }
        \toprule
        \rowcolor{Gray} \textbf{Model} & \multicolumn{2}{c}{\textbf{WER \%}} \\
        \cline{2-3}
        \rule{0pt}{3ex} & \textbf{ATCO test} & \textbf{Pilot test}  \\
        \midrule 
        Clean & 36.9 & 47.7 \\
        Noise & \bftab 31.3 & \bftab 41.1 \\
        Combined & 36.1 & 45.8 \\
        Multitask &  31.6 & \bftab 41.1 \\
        \bottomrule
    \end{tabular}}
    \vspace{2mm}
\end{table}  

\section{Experiments}
\label{sec:experiments}

For all our experiments, conventional biphone Convolutional Neural Network (CNN)~\cite{lecun1995convolutional} + TDNN-F~\cite{povey2018semi} based acoustic models trained with Kaldi~\cite{povey2011kaldi} toolkit (i.e., nnet3 model architecture) is used. AMs are trained with the LF-MMI~\cite{povey2016purely} training framework, considered to produce state-of-the-art performance for hybrid ASR systems. In all the experiments, 3-fold speed perturbation~\cite{ko2015audio} and i-vectors features are used. The multitask training script used can be found in Kaldi~\cite{povey2011kaldi}.\footnote{The script in located in: \url{https://github.com/kaldi-asr/kaldi/blob/master/egs/babel/s5d/local/chain2/run_tdnn.sh}.} The value of the task dependent weight $\langweight$ used in our experiments is $0.5$.
Language model (LM) is trained with all the manual transcripts available from datasets described in Section~\ref{OthData} and used for all the experiments.

The performance of different models is evaluated on LiveATC test set with the Word Error Rate (WER) metric. WER is computed with the Levenshtein distance at the word level. The total duration of the test set is 1h~50~mins. The set is split into two subsets: ATCO set (52~mins) and Pilot set (58~mins). In each group of experiments, results are given for i)~AM trained for each task separately, ii)~AM trained by combining all data and iii)~AM trained with multitask learning.

\begin{table}[t]
    \caption{\textbf{Comparison of word error rates (WER) in percentage for acoustic models trained with only the data collected from VHF receivers}. The models are tested on LiveATC ATCO and pilot test sets.}
    \label{tab:mt2}
    \centering
    \scalebox{1.2}{
    \begin{tabular}{ ccc }
        \toprule
        \rowcolor{Gray} \textbf{Model} & \multicolumn{2}{c}{\textbf{WER \%}} \\
        \cline{2-3}
        \rule{0pt}{3ex} & ATCO test & Pilot test  \\
        \midrule 
        VHF ATCO & 43.2 & 51.6 \\
        VHF Pilot & 40.3 &  45 \\
        Combined & 46 & 50 \\
        Multitask & \bftab 38.2 & \bftab 44 \\
        \bottomrule
    \end{tabular}}
\end{table}

\subsection{Experiments on ATC Databases}
\label{sec:atc-db}

In this setup, we use data from the ATC databases mentioned in Section~\ref{OthData} as Clean data and its noise augmented part as Noise data. In this setup, the data is not split to ATCO and pilot. As shown in Table~\ref{tab:mt1}, both ATCO and pilot test sets provide better performance when the model is trained with Noise data compared to the model trained with only Clean data. This shows that the noise augmented version of the clean data matches with the test sets much better than the clean version.
Moreover, the Combined system performs significantly worse than the Noise system.
This shows that using the Clean dataset in fact hurts ASR performance.
This is one of the reasons why the multitask system performs only on par with the Noise system. Thus, only the noise augmented data is used  for training in the next experiments.

\subsection{Experiments on VHF Data}
\label{sec:exp-vhf}

Results in Table~\ref{tab:mt2} are presented for AMs trained with only the VHF data. Applying speaker role identification for the pre-processed data (66~h) yields 43~h for ATCO and 23~h for Pilot. 
Similar to Table~\ref{tab:mt1}, the results in Table~\ref{tab:mt2} show that using multitask learning instead of training AM by combining all the data provides better ASR performance. Furthermore, the results reveal that due to the low amount of data, multitask learning outperforms its single task counterparts. 

\begin{table}[t]
    \caption{\textbf{Comparison of word error rates (WER) in percentages for acoustic models trained with all ATCO and pilot data from all databases}. Additionally, the training data is augmented with noise.}
    \label{tab:mt3}
    \centering
    \scalebox{1.2}{
    \begin{tabular}{ ccc }
        \toprule
        \rowcolor{Gray} \textbf{Model} & \multicolumn{2}{c}{\textbf{WER \%}} \\
        \cline{2-3}
        \rule{0pt}{3ex} & ATCO test & Pilot test  \\
        \midrule
        ATCO & \bftab 30.3 & 43.2 \\
        Pilot & 32.8 & \bftab 40.3 \\
        Combined & 31.2 & 41.3 \\
        Multitask &  31.9 & 41.3 \\
        \bottomrule
    \end{tabular}}
\end{table} 

\subsection{Experiments on VHF and Other ATC Datasets}
\label{sec:exp-8db}

In this subsection, we report results with models trained from  both VHF and ATC datasets used in the previous two experiments. By investigating the ATC databases used in Section~\ref{sec:atc-db}, we discovered that some datasets also contain pilot speech. 
Since no speaker role labels are available for these sets, we applied the proposed method to split the noise augmented speech as ATCO or pilot and combined them with their respective classes of the VHF data. This provided 123h of data for ATCO and 80h for pilot. The results in Table~\ref{tab:mt3} show that training AMs for each task separately performs relatively better, by 2.9\% for ATCO and 2.4\% for pilot, than using the Combined system. This suggests that when more data is available, using our grammar-based approach to obtain speaker role information to train separate ATCO and pilot ASR is better than the Combined approach. The multitask system does not perform better than the Combined one. This means that there is a negative transfer when considering ATCO and pilot tasks. This is expected as the ATC data dominates in size during training.

\section{Conclusions and Future Work}

In  this  work,  we  compared  different  types  of  training  AMs with state-of-the-art  LF-MMI framework for ATCO and pilot speech recordings. The developed ASR systems were evaluated separately on ATCO and pilot test sets built from LiveATC.  Due to the noisy nature of both ATCO and pilot test sets, AM trained with only noise augmented speech data boosts the ASR performance. We proposed a simple grammar based approach to identify speaker roles automatically and train acoustic models either by speaker role or in a multitask fashion. The results show that multitask training approach outperforms  other training methods when limited training data is available. When sufficient data is available, we show that training AMs separately provides better ASR performance for both ATCO and pilot compared to the model trained by combining all data. Relative improvements of 3.2\% for the ATCO set and 1.9\% for the pilot set were obtained.

As mentioned earlier, the rule-based approach can further be improved by taking into account all the allowed variants of a callsign and using the context prior to the callsigns during classification. In our current work, we explored only the acoustic modeling part of speech recognizer. As a part of our future work, we consider investigating the improvement of  speaker-dependent ASR systems by i) training a separate LM for each speaker class or ii) interpolating the class specific LM with the baseline LM. 

\section*{Acknowledgements}
The work was supported by SESAR EC project No. 884287\,-\,HAAWAII (Highly automated air-traffic controller workstations with artificial intelligence integration). The work was also partially supported by the European Union's Horizon 2020 project No. 864702\,-\,ATCO2 (Automatic collection and processing of voice data from air-traffic communications), which is a part of Clean Sky Joint Undertaking. We wish to acknowledge Santosh Kesiraju for providing valuable insights and suggestions regarding the assignment of scores for classification.

\bibliographystyle{IEEEtran}
\bibliography{biblio}

\end{document}